\documentclass{article}

\usepackage[preprint]{neurips_2026}

\usepackage[utf8]{inputenc} % allow utf-8 input
\usepackage[T1]{fontenc}    % use 8-bit T1 fonts
\usepackage{hyperref}       % hyperlinks
\usepackage{url}            % simple URL typesetting
\usepackage{booktabs}       % professional-quality tables
\usepackage{amsfonts}       % blackboard math symbols
\usepackage{nicefrac}       % compact symbols for 1/2, etc.
\usepackage{microtype}      % microtypography
\usepackage{xcolor}         % colors
\usepackage{wrapfig}
\usepackage{graphicx} 
\usepackage{amsmath}
\usepackage{subcaption}
\usepackage{caption}
\usepackage{amssymb}
\usepackage{algorithm}
\usepackage{algpseudocode}
\usepackage{array} 

\title{Teacher-Feature Drifting: One-Step Diffusion Distillation with Pretrained Diffusion Representations}

\author{
\textbf{Yuan Zhang}$^{1,*}$ \quad
\textbf{Chenyi Li}$^{2,*}$ \quad
\textbf{Haodong Yu}$^{3,*}$ \quad
\textbf{Guoqing Ma}$^{1,\spadesuit}$ \\
\textbf{Jiajun Zha}$^{5}$ \quad
\textbf{Yuanming Yang}$^{4}$ \quad
\textbf{Bo Wang}$^{6}$ \quad
\textbf{Wei Tang}$^{1}$ \\
\textbf{Wenbo Li}$^{1}$ \quad
\textbf{Haoyang Huang}$^{1}$ \quad
\textbf{Nan Duan}$^{1}$ \\[0.6em]
{\small
$^{1}$JD Explore Academy, China \quad
$^{2}$Peking University \quad
$^{3}$Harbin Institute of Technology
} \\
{\small
$^{4}$Tsinghua University \quad
$^{5}$The Hong Kong University of Science and Technology
} \\
{\small
$^{6}$Beijing Institute of Technology
} \\[0.4em]
{\small
$^{*}$Equal contribution.
\qquad
$\spadesuit$ Project Leader.
}
}

\begin{document}

\maketitle

\makeatletter
\begingroup
\renewcommand{\thefootnote}{}
\renewcommand{\@makefntext}[1]{\noindent #1}

\footnotetext{%
\footnotesize
\textbf{Contact:}
Yuan Zhang: \href{mailto:zhangyuan.430@jd.com}{zhangyuan.430@jd.com};
Chenyi Li: \href{mailto:lichenyi@stu.pku.edu.cn}{lichenyi@stu.pku.edu.cn};
Haodong Yu: \href{mailto:24s103334@stu.hit.edu.cn}{24s103334@stu.hit.edu.cn};
Nan Duan: \href{mailto:duannan@jd.com}{duannan@jd.com}.
\quad
\textbf{Code:}
\url{https://github.com/chenyili0818/Teacher_feature_drifting}
}

\endgroup
\makeatother

\begin{abstract}
Sampling from pretrained diffusion and flow-matching models typically requires many forward passes to generate diverse and high-fidelity images. Existing distillation methods often rely on multiple auxiliary networks, carefully designed training stages, or complex optimization pipelines. In this work, we revisit the recently proposed Drifting Model objective and show that a single drifting loss can be directly used to simplify one step  distillation. A key observation is that the pretrained diffusion teacher itself already provides a strong representation space. Unlike the original Drifting Model, which relies on an additional pretrained feature extractor, we use intermediate hidden states of the pretrained teacher model as the feature representation. This removes the need for training or introducing an extra representation network while preserving a semantically meaningful feature geometry for drifting. Furthermore, we introduce a lightweight mode coverage loss to mitigate mode collapse during distillation and encourage the student generator to cover diverse teacher-supported regions. Extensive experiments on ImageNet and SDXL demonstrate that our method achieves efficient one step  generation with competitive image quality and diversity, achieving FID scores of 1.58 on ImageNet-64$\times$64 and 18.4 on SDXL, while substantially simplifying the overall distillation framework.
% Abstract

\end{abstract}

\section{Introduction}

Generative models~\citep{ho2020denoising,lipman2022flow,song2020score,liu2022flow} based on iterative denoising or transport steps have achieved remarkable progress in high-fidelity image generation. However, their strong performance typically comes at the cost of expensive inference, since producing a single image often requires tens or hundreds of neural network evaluations. This computational burden becomes particularly severe for high-resolution text-to-image synthesis, where large-scale diffusion models repeatedly evaluate heavy denoising networks. To alleviate this limitation, recent research has explored few-step and one step  generative models that aim to retain the quality of pretrained diffusion or flow-matching models while substantially reducing sampling cost.

A common strategy for accelerating diffusion and flow-based models is distillation, where a powerful multi-step teacher is compressed into a fast student generator. Existing distillation methods have achieved impressive results, but often at the cost of additional training components or carefully engineered procedures. For example, Distribution Matching Distillation (DMD)~\citep{yin2024one} and its successor DMD2~\citep{yin2024improved} rely on auxiliary fake-score networks, adversarial discriminators, or multi-stage training schedules. These components improve generation quality, but also make the overall pipeline more complex and less scalable. Another line of work trains one step  generators through trajectory-level constraints, such as Consistency Models~\citep{song2023improved,song2023consistency} and MeanFlow~\citep{geng2025mean}. However, these methods may suffer from unstable training or require Jacobian-vector products (JVPs), increasing training complexity. These limitations motivate a simpler distillation framework that directly trains a one step  generator with a single objective.

In this work, we revisit the recently proposed Drifting Model~\citep{deng2026generative} objective as a simple alternative for one step  diffusion distillation. Drifting provides a direct attraction-repulsion loss for training one step  generators, but its effectiveness depends heavily on the representation space where sample distances are measured. The original Drifting Model therefore relies on an external pretrained feature encoder, such as a self-supervised model or a latent-MAE, to compute the drifting field. We instead use the pretrained diffusion teacher itself as the drifting feature encoder. Specifically, we extract intermediate hidden states from the frozen teacher and compute the drifting loss in this teacher-induced representation space. This aligns the distillation objective with the model being distilled and removes the need for a separate representation network. We further find that extracting teacher features from moderately noised inputs, rather than clean samples, consistently improves FID, suggesting that noise smooths the teacher feature geometry and suppresses high-frequency discrepancies. In addition, we introduce a lightweight anchor-margin loss to encourage coverage of teacher supported feature regions, thereby mitigating mode collapse during diffusion distillation. Our main contributions are summarized as follows:
\begin{itemize}
    \item We propose \emph{Teacher-Feature Drifting} (TFD), a simple one step  diffusion distillation framework that computes the drifting objective directly in the hidden representation space of the pretrained teacher. By using the teacher model itself as the feature extractor, TFD avoids an additional representation network. We further show that extracting teacher features from moderately noised inputs improves the feature geometry for distillation.

    \item We introduce a lightweight anchor-margin loss to improve sample diversity after distillation. The loss encourages generated samples to cover teacher-supported feature regions, providing a complementary coverage signal to the local attraction-repulsion drifting objective and mitigating mode collapse.

    \item We conduct extensive experiments on ImageNet-$64\times64$ and SDXL text-to-image generation. TFD achieves efficient one step  generation with competitive image quality and diversity, reaching FID scores of $1.58$ on ImageNet-$64\times64$ and $18.4$ on SDXL, while substantially simplifying the overall distillation pipeline.
\end{itemize}

\begin{figure}[htbp]
    \centering
    \includegraphics[width=\linewidth]{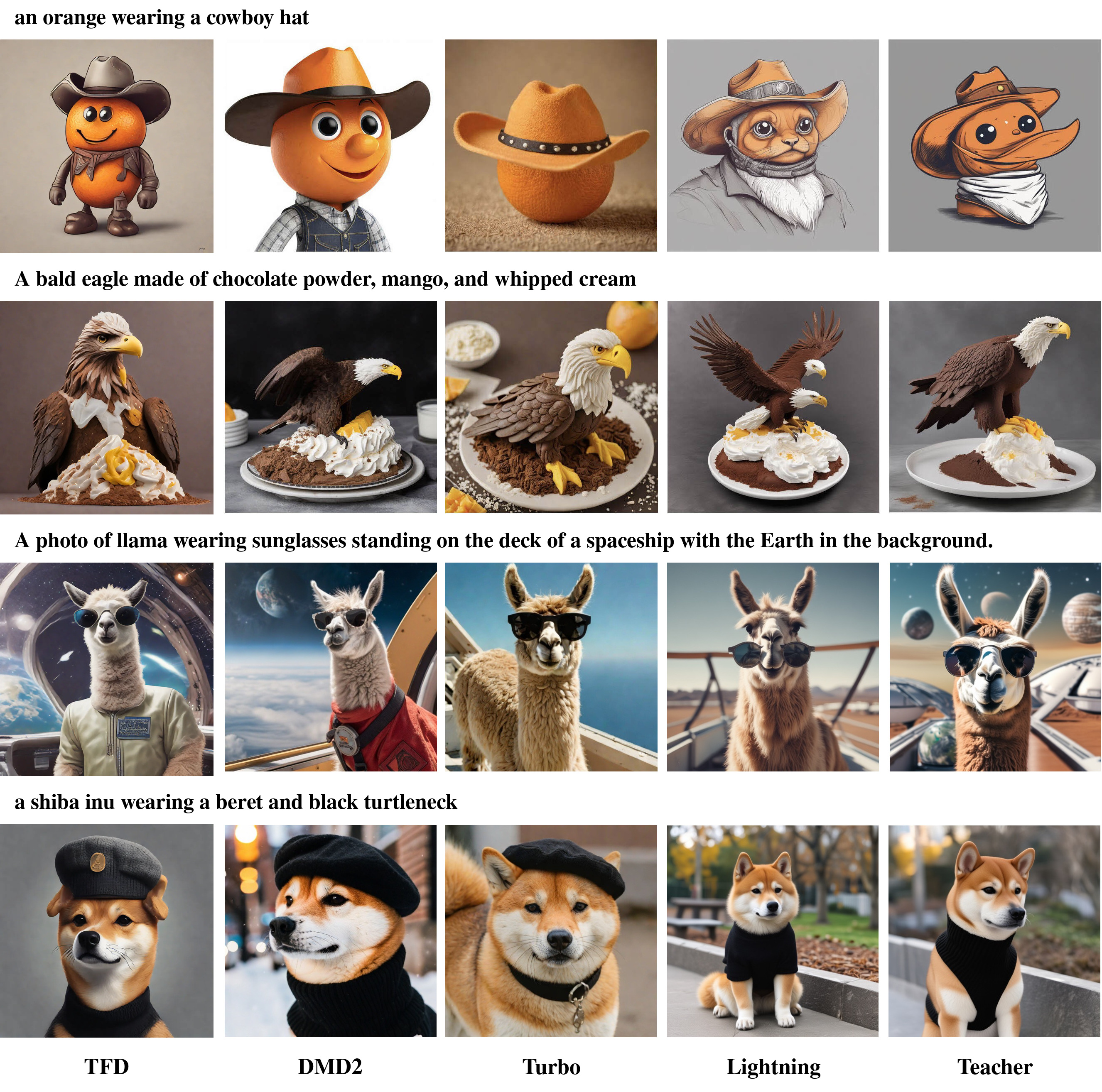}
    \caption{Qualitative comparison on SDXL text-to-image generation. TFD generates images in 
    \textbf{one step}, while the other distilled baselines use \textbf{four steps} and the SDXL teacher uses 50 steps with classifier-free guidance. Across diverse prompts, TFD preserves coherent structure, visual detail, and prompt alignment despite using substantially fewer sampling steps.}
    \label{fig:sdxl_qualitative}
\end{figure}

\section{Related Works}

\paragraph{Diffusion Distillation.}
Recent diffusion distillation methods seek to compress pretrained diffusion models into one step  or few-step generators while retaining high sample quality. Distribution Matching Distillation (DMD) formulates one step  diffusion distillation as a distribution-level matching problem, training a student generator to match the teacher distribution rather than explicitly imitating teacher sampling trajectories~\citep{yin2024one}. Its objective combines a distribution matching term, estimated through score-function differences, with a regression loss on teacher-generated noise--image pairs for training stability. DMD2 further improves this framework by eliminating the costly regression dataset, adopting a two-time-scale update for more accurate score estimation, incorporating adversarial supervision, and extending the method to multi-step student sampling~\citep{yin2024improved}. Closely related works explore alternative objectives for efficient diffusion distillation: ADD and LADD combine diffusion supervision with adversarial training for low-step high-fidelity synthesis~\citep{sauer2024adversarial,sauer2024fast}, while SiD, EMD, and SIM investigate score-identity, maximum-likelihood, and score-implicit matching formulations for one step  generation~\citep{zhou2024score,xie2024distillation,luo2024score}. Overall, these methods mark a shift from trajectory-level imitation toward distributional, adversarial, and score-based objectives for fast diffusion generation.

\paragraph{Drifting Models.}
Recent work on drifting models provides a new perspective on one step  generative modeling by moving the evolution of the pushforward distribution from inference time to training time~\citep{deng2026generative}. Instead of solving a reverse SDE/ODE as in diffusion and flow-based models, drifting models learn a single-pass generator whose samples are updated during training by a kernel-induced drifting field. Concurrent theoretical studies further clarify the role of drifting. \citet{lai2026unified} show that kernel mean-shift drifting can be interpreted as a score-matching-style objective on smoothed distributions. Similarly, \citet{turan2026generative} reveal that, under Gaussian kernels, the drift operator corresponds to a score difference between smoothed distributions, and further analyze the spectral convergence behavior and the role of stop-gradient in drifting dynamics. Complementarily, \citet{he2026sinkhorndrifting} connect drifting dynamics to Sinkhorn-divergence gradient flows and propose a Sinkhorn-normalized variant that improves stability.

\paragraph{Diffusion Representations.}
Beyond their role as generative models, pretrained diffusion models can also be used as effective feature extractors. Despite being trained through denoising, their intermediate representations capture semantic and discriminative cues that transfer to tasks such as classification, segmentation, correspondence, and visual recognition~\citep{mukhopadhyay2024textfree,xiang2023denoising,fuest2024diffusion, zhang2025generative}. This has inspired hybrid perception--generation models that jointly support recognition and diffusion-based synthesis~\citep{yang2022diffusion,deja2023diffusion,tian2024generative}, as well as distillation methods that transfer diffusion-model knowledge into specialized representation networks~\citep{yang2023repfusion,li2023dreamteacher}. Related representation-alignment methods such as REPA further demonstrate that the quality and geometry of visual representations can substantially affect diffusion and flow-based generative training~\citep{yu2025repa,leng2025repae,wang2025repa,singh2025what}. These works collectively suggest that diffusion models induce feature spaces with useful semantic structure, making them valuable not only for generation but also as representation backbones or feature metrics.

\begin{algorithm}[t]
\caption{Teacher Feature Drifting (TFD)}
\label{alg:tfd}
\begin{algorithmic}[1]
\Require Dataset $\mathcal{D}$, one step generator $f_\theta$, frozen teacher $T$, layers $\mathcal{S}$, anchor banks $\{\mathcal{A}^\ell\}$.
\Repeat
    \State Sample reference batch $\{(x_i^+,c_i)\}_{i=1}^{B}\sim p_{\rm ref}$ and noise $\epsilon_i\sim p_\epsilon$, where $p_{\rm ref}$ can be either the data distribution or the teacher-generated distribution.
    \State Generate samples $x_i^- = f_\theta(\epsilon_i,c_i)$.

    \For{$\ell\in\mathcal{S}$}
        \State Extract real and generated teacher features:
        \[
        r_i^\ell=h_T^\ell(x_i^+ + \sigma_{t_f}\xi_i^+,t_f,c_i),
        \qquad
        z_i^\ell=h_T^\ell(x_i^- + \sigma_{t_f}\xi_i^-,t_f,c_i).
        \]
        \State Estimate $V_{p_T^\ell,q_T^\ell}(z_i^\ell)$ by the mini-batch version of Eq.~\eqref{eq:drift_field}.
        \State Accumulate $\mathcal{L}_{\rm TFD}$ using Eq.~\eqref{eq:tfd_loss}.
        \State Sample anchors $\{a_m^\ell\}_{m=1}^{M}\sim\mathcal{A}^\ell$.
        \State Compute $s_m^\ell$, $\bar{s}_m^\ell$, and accumulate $\mathcal{L}_{\rm anchor}$ using Eqs.~\eqref{eq:anchor_gen_support}--\eqref{eq:anchor_margin_loss}.
    \EndFor

    \State Update $\theta$ by minimizing Eq.~\eqref{eq:final_loss}.
\Until{convergence}
\State \Return $f_\theta$
\end{algorithmic}
\end{algorithm}

\section{Teacher Feature Drifting}
\label{sec:teacher_feature_drifting}

\subsection{Preliminaries: Drifting Models}

Let $f_\theta$ be a one step  generator that maps a noise variable
$\epsilon \sim p_\epsilon$ to a generated sample
\begin{equation}
    x = f_\theta(\epsilon), \qquad x \sim q_\theta,
\end{equation}
where $q_\theta = (f_\theta)_\# p_\epsilon$ denotes the pushforward distribution induced by $f_\theta$.
The goal of generative modeling is to learn $f_\theta$ such that $q_\theta$ matches the target distribution
$p_{\rm data}$.

Drifting models formulate one step  generation as the evolution of the pushforward distribution during training.
Instead of simulating an iterative denoising trajectory at inference time, drifting models define a distribution-dependent
field that moves current generated samples toward the data distribution. For simplicity, we write all quantities in a
representation space where the kernel distance is computed. This representation can be the sample space itself or a
feature space induced by a frozen encoder.

Given a point $x$, a data distribution $p$, and the current model distribution $q$, drifting models define an
attraction field from real samples and a repulsion field from generated samples:
\begin{equation}
    V_p^+(x)
    =
    \frac{
    \mathbb{E}_{y^+ \sim p}
    \left[
    k(x,y^+) (y^+ - x)
    \right]
    }{
    \mathbb{E}_{y^+ \sim p}
    \left[
    k(x,y^+)
    \right]
    },
    \qquad
    V_q^-(x)
    =
    \frac{
    \mathbb{E}_{y^- \sim q}
    \left[
    k(x,y^-) (y^- - x)
    \right]
    }{
    \mathbb{E}_{y^- \sim q}
    \left[
    k(x,y^-)
    \right]
    } .
\end{equation}
Here $k(\cdot,\cdot)$ is a similarity kernel, typically chosen as a Laplace kernel,
\begin{equation}
    k(x,y) = \exp\left(-\frac{\|x-y\|_2}{\tau}\right),
\end{equation}
where $\tau$ is a temperature parameter. The normalized kernel weights make each term a local mean-shift direction:
$V_p^+(x)$ pulls $x$ toward nearby real samples, while $V_q^-(x)$ pulls $x$ toward nearby generated samples.
The final drifting field is defined as their difference:
\begin{equation}
    V_{p,q}(x)
    =
    \left(
    V_p^+(x) - V_q^-(x)
    \right),
    \label{eq:drift_field}
\end{equation}
when $p=q$, the two
mean-shift fields coincide and the drifting field vanishes, yielding the equilibrium condition
\begin{equation}
    p=q \quad \Rightarrow \quad V_{p,q}(x)=0 .
\end{equation}

The generator is optimized with a fixed-point regression objective. For a generated sample $x=f_\theta(\epsilon)$,
we first construct a frozen transported target
\begin{equation}
    \tilde{x}
    =
    x + V_{p_{\rm data},q_\theta}(x),
\end{equation}
and then regress the generator output toward this target:
\begin{equation}
    \mathcal{L}_{\rm drift}
    =
    \mathbb{E}_{\epsilon \sim p_\epsilon}
    \left[
    \left\|
    f_\theta(\epsilon)
    -
    \operatorname{sg}
    \left(
    f_\theta(\epsilon)
    +
    V_{p_{\rm data},q_\theta}
    \left(
    f_\theta(\epsilon)
    \right)
    \right)
    \right\|_2^2
    \right],
    \label{eq:drift_loss}
\end{equation}
where $\operatorname{sg}(\cdot)$ denotes stop-gradient. 
A crucial practical component of drifting models is the choice of representation used to compute the kernel distance. For high-dimensional image generation, computing $k(x,y)$ directly in pixel or latent space often fails to capture semantic similarity. Therefore, prior drifting models use an additional pretrained representation encoder, such as a self-supervised encoder or a latent-MAE, and compute the above field in that feature space. In this work, we remove this extra representation network by using intermediate hidden states of the pretrained diffusion teacher as the drifting representation.

\subsection{Teacher Hidden States as the Drifting Space}

We use the internal representations of the pretrained diffusion teacher as the space where the drifting loss is computed.
Let $T$ denote a frozen teacher model, and let $h_T^\ell(\cdot)$ be the hidden activation from layer $\ell$.
Given a generated or real sample $x$ with condition $c$, a straightforward choice is to directly feed the clean sample into the teacher and extract its intermediate feature:
\begin{equation}
    \phi_{T,0}^\ell(x,c)
    =
    h_T^\ell(x, t_f, c),
\end{equation}
where $t_f$ denotes the teacher timestep used for feature extraction. We find that these clean teacher features already provide a useful representation for drifting: they contain semantic and structural information learned by the pretrained teacher, and can be used to compare real and generated samples without introducing a separate representation encoder.

Furthermore, we empirically observe that extracting features from moderately noised samples leads to better distillation performance. Specifically, before feeding $x$ into the teacher, we perturb it with the teacher's forward noising process:
\begin{equation}
    x_{t_f}
    =
     x + \sigma_{t_f}\xi,
    \qquad
    \xi \sim \mathcal{N}(0,I),
\end{equation}
and define the teacher feature as
\begin{equation}
    \phi_T^\ell(x,c)
    =
    h_T^\ell(x_{t_f}, t_f, c).
\end{equation}
In practice, a moderate noise level, e.g., $\sigma_{t_f}=0.1$, consistently improves the final FID compared with clean-feature extraction. We hypothesize that such perturbation smooths the teacher representation, suppressing high-frequency artifacts and making the feature geometry more suitable for distribution-level matching. Given teacher features from selected layers $\mathcal{S}$, we apply the drifting objective introduced in the previous subsection directly in this teacher-induced feature space:
\begin{equation}
    \mathcal{L}_{\rm TFD}
    =
    \sum_{\ell \in \mathcal{S}}
    \mathbb{E}_{\epsilon,c}
    \left[
    \left\|
    \phi_T^\ell(x,c)
    -
    \operatorname{sg}
    \left(
    \phi_T^\ell(x,c)
    +
     V_{p_T^\ell,q_T^\ell}
    \left(
    \phi_T^\ell(x,c)
    \right)
    \right)
    \right\|_2^2
    \right],
    \label{eq:tfd_loss}
\end{equation}
where $x=f_\theta(\epsilon,c)$, $p_T^\ell$ and $q_T^\ell$ denote the real and generated distributions in the $\ell$-th teacher layer.

\subsection{Anchor-Margin Regularization}

The drifting loss provides a local attraction-repulsion direction for each generated sample. However, since this direction is estimated from mini-batch kernel neighborhoods, some regions in the teacher feature space may receive weak support during training. To complement the drifting objective, we introduce a lightweight anchor-margin regularizer that encourages generated samples to occupy teacher-supported feature regions.

Let $\{a_i\}_{i=1}^{M}$ denote anchor features extracted by the teacher, and let $\{z_j\}_{j=1}^{N}$ denote generated features from the same teacher layer. The anchors can be obtained from real samples or teacher-generated samples, and are treated as fixed targets. For each anchor $a_i$, we estimate its support from generated samples by a kernel density estimate:
\begin{equation}
    s_i
    =
    \frac{1}{M}
    \sum_{j=1}^{M}
    \exp
    \left(
    -\frac{\|a_i-z_j\|_2}{h}
    \right),
    \label{eq:anchor_gen_support}
\end{equation}
where $h$ is the kernel bandwidth. A larger $s_i$ indicates that the anchor region is better occupied by generated samples in the teacher feature space. Rather than enforcing a fixed support threshold for all anchors, we adapt the target threshold according to the local density of the anchors themselves. Specifically, we compute the self-support of each anchor:
\begin{equation}
    \bar{s}_i
    =
    \frac{1}{M-1}
    \sum_{k \neq i}
    \exp
    \left(
    -\frac{\|a_i-a_k\|_2}{2h}
    \right),
    \label{eq:anchor_self_support}
\end{equation}
and define the desired support level as $\rho_i = \alpha \bar{s}_i$,
where $\alpha$ controls the margin strength. This adaptive threshold accounts for the non-uniform density of the anchor distribution: anchors in dense regions require higher generated support, while isolated anchors are assigned a lower target. The anchor-margin loss penalizes only anchors whose generated support falls below the desired level:
\begin{equation}
    \mathcal{L}_{\rm anchor}
    =
    \frac{1}{M}
    \sum_{i=1}^{M}
    \max(0, \rho_i - s_i).
    \label{eq:anchor_margin_loss}
\end{equation}
Thus, well-supported anchors do not dominate the optimization, while under-supported anchor regions continue to provide gradients to nearby generated samples. The final objective is
\begin{equation}
    \mathcal{L}
    =
    \mathcal{L}_{\rm TFD}
    +
    \lambda_{\rm anchor}
    \mathcal{L}_{\rm anchor},
    \label{eq:final_loss}
\end{equation}
where $\lambda_{\rm anchor}$ balances the drifting loss and the anchor-margin regularizer. Since both terms are computed in the same teacher feature space, this regularization introduces no additional networks and adds only a small computational overhead.

\section{Experiments}

\subsection{ImageNet-64$\times$64}

%% Anchor Margin Loss

\paragraph{Setting.}
We evaluate our method on class-conditional ImageNet \citep{deng2009imagenet} generation at $64\times64$ resolution. We compare our algorithm with DMD~\citep{yin2024one} and DMD2~\citep{yin2024improved} under the same pretrained class-conditional EDM teacher~\citep{karras2022elucidating}. We follow the standard ImageNet-$64 \times 64$
evaluation protocol in this line of work. Each model generates $50{,}000$ samples, class labels are
balanced over the $1000$ ImageNet classes, and sample quality is measured by FID \citep{heusel2017gans}.  The student is a one step  class-conditional generator initialized from the pretrained EDM teacher. The teacher is kept frozen throughout training and is used only to provide feature representations for the training objective. More training details are given in Appendix \ref{appendix: hyperparameters}.

\paragraph{Feature Extraction.} We compute the drifting objective over five teacher feature levels: the 6th and 11th encoder blocks, the bottleneck block, the 7th and the 12th decoder blocks. For teacher feature extraction, we sample the feature noise level at $0.1$.

\begin{figure}[t]
    \centering
    \begin{minipage}[t]{0.47\textwidth}
        \vspace{0pt}
        \centering
        \small
        \setlength{\tabcolsep}{3pt}
        \begin{tabular}{lcc}
            \toprule
            Method & FP & FID $\downarrow$ \\
            \midrule
            EDM teacher (ODE)~\citep{karras2022elucidating} & 511 & 2.32 \\
            EDM teacher (SDE)~\citep{karras2022elucidating} & 511 & 1.36 \\
            \midrule
            Consistency model~\citep{song2023consistencymodels} & 1 & 6.20 \\ 
            iCT-deep~\citep{song2023improved} & 1 & 3.25 \\
            DMD~\citep{yin2024one} & 1 & 2.62 \\
            EMD~\citep{xie2024distillation} & 1 & 2.20 \\
            CTM~\citep{kim2023consistency} & 1 & 1.92 \\
            DMD2 (200K)~\citep{yin2024improved} & 1 & \textbf{1.51} \\
            \midrule
            TFD (Ours) (200K) & 1 & \underline{1.58} \\
            \bottomrule
        \end{tabular}
        \captionof{table}{Class conditional ImageNet-$64 \times 64$ generation with EDM-based one step  generators. FP denotes the number of forward passes. \textbf{Bold} and \underline{underlined} indicate the best and second-best results.}
        \label{tab:imagenet64-edm-onestep}
    \end{minipage}
    \hfill
    \begin{minipage}[t]{0.48\textwidth}
        \vspace{0pt}
        \centering
        \includegraphics[width=\linewidth]{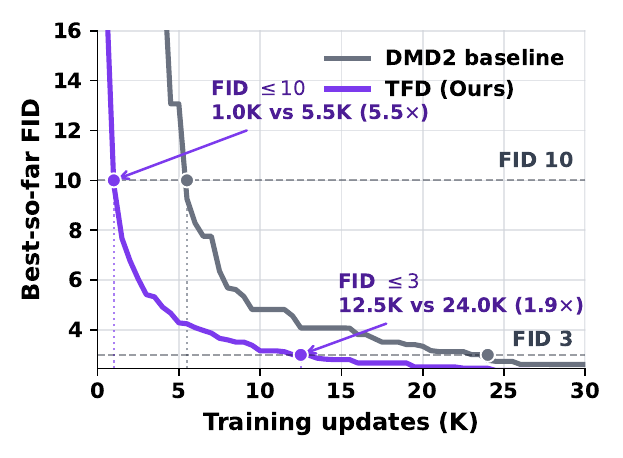}
        \caption{Training acceleration comparison. Our method reaches both FID $\leq 10$ and FID $\leq 3$ substantially earlier than DMD2, indicating faster convergence for one step  generation.}
        \label{fig:acceleration}
    \end{minipage}
\end{figure}

\paragraph{Performance.}
Table~\ref{tab:imagenet64-edm-onestep} summarizes the class-conditional ImageNet-$64 \times 64$ results under the EDM teacher. TFD achieves an FID of $1.58$ with a single forward pass after $200$k training iterations. This result substantially improves over representative one step  distillation baselines, including DMD, EMD, and CTM, and is close to the full DMD2 model under a comparable training budget. Unlike DMD-style approaches that rely on auxiliary distribution-matching components, TFD does not train an additional fake-score network or adversarial classifier. Instead, the student is optimized directly through drifting in the frozen teacher's internal representation space, together with an anchor-margin coverage term that encourages generated samples to cover data-supported regions.

Figure~\ref{fig:acceleration} further shows that TFD provides a more efficient optimization trajectory. In the early stage of training, TFD reaches FID $\leq 10$ after $1.0$k updates, whereas the DMD2 baseline requires $5.5$k updates to reach the same level. The advantage remains at a stricter quality threshold: TFD reaches FID $\leq 3$ after $12.5$k updates, compared with $24.0$k updates for DMD2. These results suggest that teacher-feature drifting supplies a strong training signal for one step  generation, while the data-side coverage term helps mitigate missing-mode behavior. 

\subsection{SDXL Text-to-Image Generation}

\paragraph{Setting.}
We evaluate our method on text-to-image generation by distilling Stable Diffusion XL (SDXL)~\citep{podell2023sdxl}. Following prior text-to-image distillation work~\citep{yin2024improved}, we evaluate zero-shot generation on COCO 2014~\citep{lin2014microsoft}.  During training, for each sampled prompt, we construct the positive set by first searching for prompt-matched real images in our paired image dataset. Let $N_+$ denote the number of positive samples required for each prompt. If at least $N_+$ matched real images are available, we use real images only. If the number of matched real images is insufficient, we use all available real images and fill the remaining positive slots with samples generated by the SDXL teacher. When no matched real image is available, all positives are sampled from the teacher. Teacher positives are generated with $50$ denoising steps and classifier-free guidance scale $6.0$. This hybrid construction allows us to exploit real prompt-image pairs whenever available, while still supporting prompts without sufficient paired images.

\paragraph{Feature Extraction.} For SDXL experiments, we compute the drifting objective using three intermediate feature levels from the frozen teacher U-Net: \texttt{down\_blocks.2}, \texttt{mid\_block}, and \texttt{up\_blocks.0}. These layers capture mid-level encoder features, bottleneck representations, and early decoder features, respectively. Unless otherwise specified, teacher features are extracted from moderately noised inputs with feature noise level $0.1$.

\paragraph{Performance.} Table~\ref{tab:sdxl-coco} reports COCO 2014 results for SDXL-based text-to-image generation. TFD achieves the best FID among the compared distilled models, reaching 18.39 with only one forward pass, while maintaining a competitive CLIP score of 0.332. Compared with prior accelerated SDXL variants, TFD improves image fidelity without sacrificing text-image alignment, suggesting that the teacher-induced feature space provides an effective supervision signal for one step  distillation.

Figure~\ref{fig:sdxl_qualitative} provides qualitative comparisons across diverse prompts. Despite using fewer sampling steps, our one step generator produces coherent objects, preserves clear visual details, and maintains prompt alignment, achieving perceptual quality comparable to four step distilled baselines and the 50 step SDXL teacher. The examples also show that TFD handles different types of prompts, including object-centric descriptions, compositional scenes, and stylized attributes. These results indicate that teacher-feature drifting can effectively distill SDXL into a fast one step  generator while preserving both visual realism and prompt consistency, without introducing a complex training pipeline.

\begin{table}
    \centering
    \captionof{table}{COCO 2014 generation with SDXL-based one step  generators. FP denotes the number of forward passes. \textbf{Bold} and \underline{underlined} indicate the best and second-best results.}
   \begin{tabular}{lccc}
            \toprule
            Method & FP & FID $\downarrow$ & CLIP $\uparrow$\\
            \midrule
            SDXL teacher (cfg=6)~\citep{podell2023sdxl} & 100 & 19.36 & 0.332 \\
            SDXL teacher (cfg=8)~\citep{podell2023sdxl}  & 100 & 20.39 & 0.335 \\
            \midrule
            LCM-SDXL~\citep{luo2023lcm} & 1 & 81.62 & 0.275 \\
            SDXL-Turbo~\citep{sauer2024adversarial} & 1 & 24.57 & \textbf{0.337} \\
            SDXL Lightning~\citep{lin2024sdxl} & 1 & 23.92 & 0.316 \\
            DMD2~\citep{yin2024improved} & 1 & \underline{19.01} & \underline{0.336} \\
            \midrule
            TFD (Ours) & 1 & \textbf{18.39} & {0.332}\\
            \bottomrule
        \end{tabular}
        \label{tab:sdxl-coco}
\end{table}

\subsection{Ablation Study}
We ablate the main design choices of TFD on class-conditional ImageNet-$64 \times 64$, including the anchor-margin coverage loss, the noise level used for teacher-feature extraction, and the set of teacher layers used as drifting representations. More ablation studies are provided in Appendix \ref{app:imagenet-ablations} and \ref{app: sdxl abl}.

\begin{figure}[ht]
    \centering
    \includegraphics[width=\linewidth]{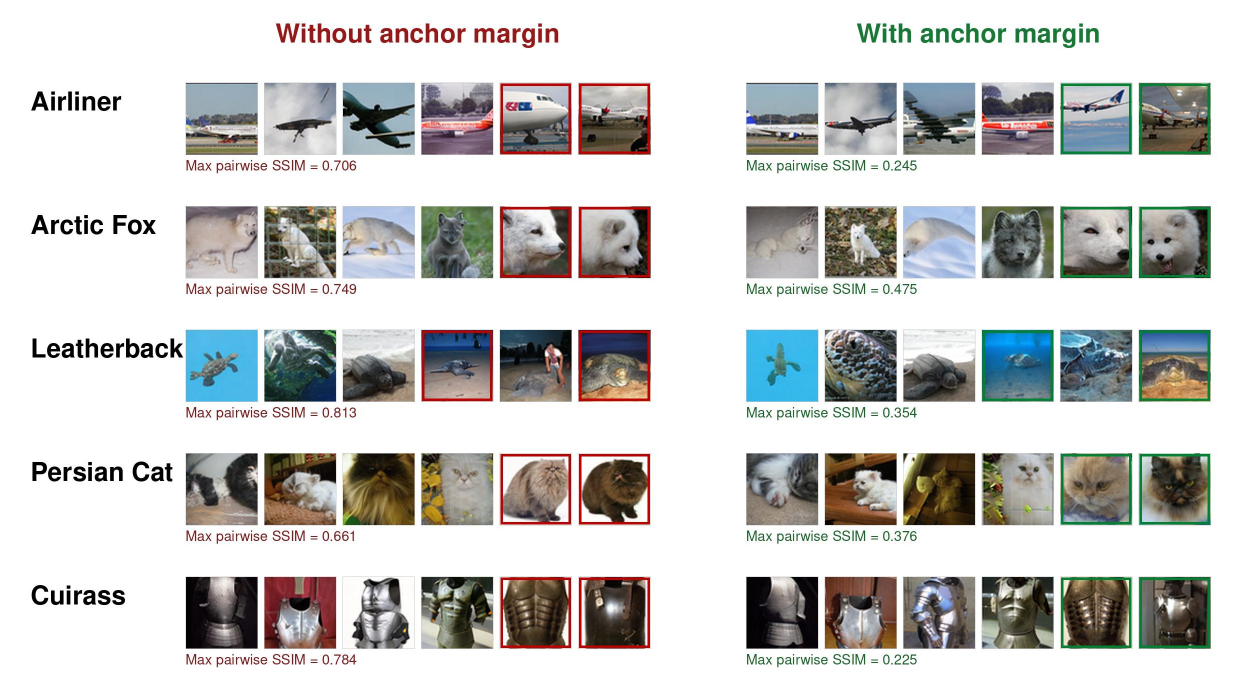}
    \caption{Effect of the anchor margin loss on sample diversity. We visualize class-conditional samples generated without and with the anchor margin loss across five ImageNet classes. Without the anchor margin, the generated samples often contain visually similar or near-duplicate images within the same class, as reflected by high maximum pairwise SSIM scores. In contrast, adding the anchor margin loss encourages samples to stay separated in the teacher feature space, reducing redundancy while preserving class semantics.}
    \label{fig:anchor-margin-ablation}
\end{figure}

\textbf{Effect of anchor-margin coverage.}
Figure~\ref{fig:anchor-margin-ablation} visualizes the effect of the anchor-margin coverage loss on ImageNet-$64\times64$. Without this term, the one step  generator tends to produce visually similar samples within the same class, leading to missing-mode behavior even when individual samples remain realistic. To quantify this effect, we report the maximum pairwise SSIM among generated samples from the same class. SSIM measures structural similarity between two images, and a higher value indicates that at least one pair of generated samples shares highly similar layouts, shapes, or textures. A lower maximum pairwise SSIM suggests reduced duplication and improved intra-class diversity. Adding the anchor-margin loss consistently lowers this metric across the illustrated classes, while also producing more diverse visual layouts. This suggests that generated-query drifting alone may provide insufficient supervision for under-covered data regions, whereas the data-side anchor constraint supplies a complementary coverage signal.

\begin{figure}
    \centering
    \includegraphics[width=\linewidth]{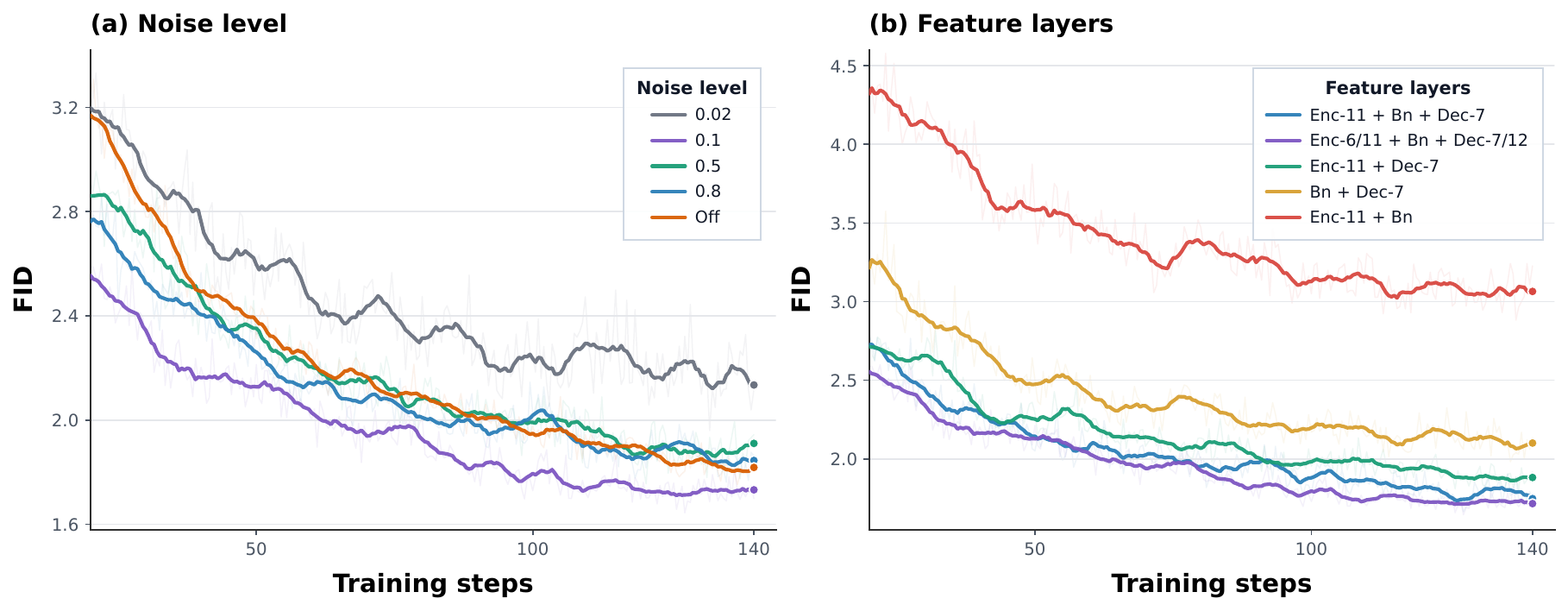}
    \caption{Ablation studies on noise level and feature layer. Enc-$k$ denotes the $k$-th encoder block, Bn denotes the bottleneck block, and Dec-$k$ denotes the $k$-th decoder block of the teacher U-Net. 
    Lower FID indicates better generation quality.}
    \label{fig:layer-noise-ablation}
\end{figure}

\textbf{Effect of teacher-feature noise level.}
We further study the noise level used when extracting features from the frozen teacher. As shown in Figure~\ref{fig:layer-noise-ablation}(a), using a small nonzero feature-noise range consistently improves training compared with either clean-feature extraction or overly narrow/broad perturbations. In particular, the noise level $0.1$ gives the best overall performance among the tested settings, reaching the lowest FID within the training budget. In contrast, using nearly fixed low noise, e.g., 0.02, converges more slowly and gives worse final FID, while larger noise ranges such as 0.5 also degrade performance. These results support the hypothesis that moderate noising smooths the teacher representation space and suppresses high-frequency discrepancies.

\textbf{Effect of teacher feature layers.}
Finally, we ablate which teacher layers are used to compute the drifting loss. Figure~\ref{fig:layer-noise-ablation}(b) shows that combining intermediate and deep teacher representations is important for stable performance. Using only a subset of deep layers, such as Bottleneck+Dec-7, performs worse than including encoder, bottleneck, and decoder features together. Similarly, using Enc-11+Bottleneck without decoder features leads to a substantial degradation, suggesting that decoder-side representations provide useful spatial and generative information for matching. The best configuration combines multiple feature levels, such as Enc-6, Enc-11, Bottleneck, Dec-7, and Dec-12, which improves final FID over the three-layer variant while maintaining stable convergence. 

\section{Conclusion}

We present Teacher-Feature Drifting (TFD), a simple one step diffusion distillation framework that applies the drifting objective directly in the internal representation space of a pretrained diffusion teacher. By using intermediate teacher hidden states as the feature space for attraction-repulsion based distribution matching, TFD removes the need for an external self-supervised feature encoder and turns the teacher itself into the metric space for distillation. This design simplifies the training pipeline while preserving a strong semantic representation geometry inherited from the pretrained teacher. Motivated by the diffusion teacher’s noisy training regime, we extract teacher features from moderately noised inputs, which better aligns feature extraction with the teacher’s native input distribution and induces a smoother drifting field for distribution matching. To mitigate mode collapse in diffusion distillation, we introduce a lightweight anchor-margin regularizer that encourages generated samples to cover teacher-supported feature regions and improves sample diversity. Experiments on ImageNet-64$\times$64 and SDXL text-to-image generation show that TFD achieves competitive one-step generation quality across different settings. Ablation studies further validate the importance of teacher-feature noise, multi-level feature selection, and anchor-margin coverage. 

\section*{Limitations}

This work has not yet explored teacher-feature drifting under Diffusion Transformer (DiT) architectures. Our current experiments focus on U-Net-based diffusion teachers, and we leave a systematic study of DiT-based teachers to future work, including how different transformer layers and representations may affect distillation performance. In addition, our evaluation is limited to the current experimental scale. We have not yet tested the method on larger-scale datasets, higher-resolution generation settings, or substantially larger teacher models. Moreover, TFD currently lacks a theoretical guidance for selecting feature layers and noise levels. Extending TFD to DiT-based models and larger-scale generation remains an important direction for future work.

\newpage
\bibliography{ref}

@article{geng2025mean,
  title={Mean flows for one-step generative modeling},
  author={Geng, Zhengyang and Deng, Mingyang and Bai, Xingjian and Kolter, J Zico and He, Kaiming},
  journal={arXiv preprint arXiv:2505.13447},
  year={2025}
}

@article{song2023consistency,
  title={Consistency models},
  author={Song, Yang and Dhariwal, Prafulla and Chen, Mark and Sutskever, Ilya},
  year={2023}
}

@article{liu2022flow,
  title={Flow straight and fast: Learning to generate and transfer data with rectified flow},
  author={Liu, Xingchao and Gong, Chengyue and Liu, Qiang},
  journal={arXiv preprint arXiv:2209.03003},
  year={2022}
}

@article{ho2020denoising,
  title={Denoising diffusion probabilistic models},
  author={Ho, Jonathan and Jain, Ajay and Abbeel, Pieter},
  journal={Advances in neural information processing systems},
  volume={33},
  pages={6840--6851},
  year={2020}
}

@article{lipman2022flow,
  title={Flow matching for generative modeling},
  author={Lipman, Yaron and Chen, Ricky TQ and Ben-Hamu, Heli and Nickel, Maximilian and Le, Matt},
  journal={arXiv preprint arXiv:2210.02747},
  year={2022}
}

@article{song2020score,
  title={Score-based generative modeling through stochastic differential equations},
  author={Song, Yang and Sohl-Dickstein, Jascha and Kingma, Diederik P and Kumar, Abhishek and Ermon, Stefano and Poole, Ben},
  journal={arXiv preprint arXiv:2011.13456},
  year={2020}
}

@inproceedings{deng2009imagenet,
  title={Imagenet: A large-scale hierarchical image database},
  author={Deng, Jia and Dong, Wei and Socher, Richard and Li, Li-Jia and Li, Kai and Fei-Fei, Li},
  booktitle={2009 IEEE conference on computer vision and pattern recognition},
  pages={248--255},
  year={2009},
  organization={Ieee}
}

@article{zhang2025generative,
  title={Generative pre-trained autoregressive diffusion transformer},
  author={Zhang, Yuan and Jiang, Jiacheng and Ma, Guoqing and Lu, Zhiying and Huang, Haoyang and Yuan, Jianlong and Duan, Nan and Jiang, Daxin},
  journal={arXiv preprint arXiv:2505.07344},
  year={2025}
}

@inproceedings{yin2024one,
  title={One-step diffusion with distribution matching distillation},
  author={Yin, Tianwei and Gharbi, Micha{\"e}l and Zhang, Richard and Shechtman, Eli and Durand, Fredo and Freeman, William T and Park, Taesung},
  booktitle={Proceedings of the IEEE/CVF conference on computer vision and pattern recognition},
  pages={6613--6623},
  year={2024}
}

@article{yin2024improved,
  title={Improved distribution matching distillation for fast image synthesis},
  author={Yin, Tianwei and Gharbi, Micha{\"e}l and Park, Taesung and Zhang, Richard and Shechtman, Eli and Durand, Fredo and Freeman, William T},
  journal={Advances in neural information processing systems},
  volume={37},
  pages={47455--47487},
  year={2024}
}

@article{karras2022elucidating,
  title={Elucidating the design space of diffusion-based generative models},
  author={Karras, Tero and Aittala, Miika and Aila, Timo and Laine, Samuli},
  journal={Advances in neural information processing systems},
  volume={35},
  pages={26565--26577},
  year={2022}
}

@article{heusel2017gans,
  title={Gans trained by a two time-scale update rule converge to a local nash equilibrium},
  author={Heusel, Martin and Ramsauer, Hubert and Unterthiner, Thomas and Nessler, Bernhard and Hochreiter, Sepp},
  journal={Advances in neural information processing systems},
  volume={30},
  year={2017}
}

@article{kim2023consistency,
  title={Consistency trajectory models: Learning probability flow ode trajectory of diffusion},
  author={Kim, Dongjun and Lai, Chieh-Hsin and Liao, Wei-Hsiang and Murata, Naoki and Takida, Yuhta and Uesaka, Toshimitsu and He, Yutong and Mitsufuji, Yuki and Ermon, Stefano},
  journal={arXiv preprint arXiv:2310.02279},
  year={2023}
}

@article{xie2024distillation,
  title={Em distillation for one-step diffusion models},
  author={Xie, Sirui and Xiao, Zhisheng and Kingma, Diederik P and Hou, Tingbo and Wu, Ying N and Murphy, Kevin and Salimans, Tim and Poole, Ben and Gao, Ruiqi},
  journal={Advances in Neural Information Processing Systems},
  volume={37},
  pages={45073--45104},
  year={2024}
}

@article{song2023improved,
  title={Improved techniques for training consistency models},
  author={Song, Yang and Dhariwal, Prafulla},
  journal={arXiv preprint arXiv:2310.14189},
  year={2023}
}

@article{deng2026generative,
  title={Generative Modeling via Drifting},
  author={Deng, Mingyang and Li, He and Li, Tianhong and Du, Yilun and He, Kaiming},
  journal={arXiv preprint arXiv:2602.04770},
  year={2026}
}

@article{lai2026unified,
  title={A Unified View of Drifting and Score-Based Models},
  author={Lai, Chieh-Hsin and Nguyen, Bac and Murata, Naoki and Takida, Yuhta and Uesaka, Toshimitsu and Mitsufuji, Yuki and Ermon, Stefano and Tao, Molei},
  journal={arXiv preprint arXiv:2603.07514},
  year={2026}
}

@article{he2026sinkhorndrifting,
  title={Sinkhorn-Drifting Generative Models},
  author={He, Ping and Khangaonkar, Om and Pirsiavash, Hamed and Bai, Yikun and Kolouri, Soheil},
  year={2026},
  journal={arXiv preprint arXiv:2603.12366},
}

@article{turan2026generative,
  title={Generative Drifting is Secretly Score Matching: a Spectral and Variational Perspective},
  author={Turan, Erkan and Ovsjanikov, Maks},
  journal={arXiv preprint arXiv:2603.09936},
  year={2026}
}

@inproceedings{sauer2024adversarial,
  title={Adversarial Diffusion Distillation},
  author={Sauer, Axel and Lorenz, Dominik and Blattmann, Andreas and Rombach, Robin},
  booktitle={European Conference on Computer Vision},
  pages={87--103},
  year={2024}
}

@inproceedings{sauer2024fast,
  title={Fast High-Resolution Image Synthesis with Latent Adversarial Diffusion Distillation},
  author={Sauer, Axel and Boesel, Frederic and Dockhorn, Tim and Blattmann, Andreas and Esser, Patrick and Rombach, Robin},
  booktitle={ACM SIGGRAPH Asia Conference Papers},
  pages={106:1--106:11},
  year={2024},
  doi={10.1145/3680528.3687625}
}

@inproceedings{zhou2024score,
  title={Score Identity Distillation: Exponentially Fast Distillation of Pretrained Diffusion Models for One-Step Generation},
  author={Zhou, Mingyuan and Zheng, Huangjie and Wang, Zhendong and Yin, Mingzhang and Huang, Hai},
  booktitle={Proceedings of the 41st International Conference on Machine Learning},
  year={2024}
}

@inproceedings{luo2024score,
  title={One-Step Diffusion Distillation through Score Implicit Matching},
  author={Luo, Weijian and Huang, Zemin and Geng, Zhengyang and Kolter, J. Zico and Qi, Guo-jun},
  booktitle={Advances in Neural Information Processing Systems},
  volume={37},
  year={2024},
  doi={10.52202/079017-3664}
}

@inproceedings{xiang2023denoising,
  title={Denoising Diffusion Autoencoders are Unified Self-supervised Learners},
  author={Xiang, Weilai and Yang, Hongyu and Huang, Di and Wang, Yunhong},
  booktitle={Proceedings of the IEEE/CVF International Conference on Computer Vision},
  pages={15802--15812},
  year={2023}
}

@inproceedings{mukhopadhyay2024textfree,
  title={Do Text-Free Diffusion Models Learn Discriminative Visual Representations?},
  author={Mukhopadhyay, Soumik and Gwilliam, Matthew and Yamaguchi, Yosuke and Agarwal, Vatsal and Padmanabhan, Namitha and Swaminathan, Archana and Zhou, Tianyi and Ohya, Jun and Shrivastava, Abhinav},
  booktitle={European Conference on Computer Vision},
  pages={253--272},
  year={2024}
}

@article{fuest2024diffusion,
  title={Diffusion Models and Representation Learning: A Survey},
  author={Fuest, Michael and Ma, Pingchuan and Gui, Ming and Schusterbauer, Johannes and Hu, Vincent Tao and Ommer, Bj{\"o}rn},
  journal={arXiv preprint arXiv:2407.00783},
  year={2024}
}

@article{yang2022diffusion,
  title={Your {ViT} is Secretly a Hybrid Discriminative-Generative Diffusion Model},
  author={Yang, Xiulong and Shih, Sheng-Min and Fu, Yinlin and Zhao, Xiaoting and Ji, Shihao},
  journal={arXiv preprint arXiv:2208.07791},
  year={2022}
}

@inproceedings{deja2023diffusion,
  title={Learning Data Representations with Joint Diffusion Models},
  author={Deja, Kamil and Trzci{\'n}ski, Tomasz and Tomczak, Jakub M.},
  booktitle={Machine Learning and Knowledge Discovery in Databases: Research Track},
  year={2023},
  publisher={Springer},
  doi={10.1007/978-3-031-43415-0_32}
}

@inproceedings{tian2024generative,
  title={{ADDP}: Learning General Representations for Image Recognition and Generation with Alternating Denoising Diffusion Process},
  author={Tian, Changyao and Tao, Chenxin and Dai, Jifeng and Li, Hao and Li, Ziheng and Lu, Lewei and Wang, Xiaogang and Li, Hongsheng and Huang, Gao and Zhu, Xizhou},
  booktitle={International Conference on Learning Representations},
  year={2024}
}

@inproceedings{yang2023repfusion,
  title={Diffusion Model as Representation Learner},
  author={Yang, Xingyi and Wang, Xinchao},
  booktitle={Proceedings of the IEEE/CVF International Conference on Computer Vision},
  pages={18938--18949},
  year={2023}
}

@inproceedings{li2023dreamteacher,
  title={DreamTeacher: Pretraining Image Backbones with Deep Generative Models},
  author={Li, Daiqing and Ling, Huan and Kar, Amlan and Acuna, David and Kim, Seung Wook and Kreis, Karsten and Torralba, Antonio and Fidler, Sanja},
  booktitle={Proceedings of the IEEE/CVF International Conference on Computer Vision},
  pages={16698--16708},
  year={2023}
}

@inproceedings{yu2025repa,
  title={Representation Alignment for Generation: Training Diffusion Transformers Is Easier Than You Think},
  author={Yu, Sihyun and Kwak, Sangkyung and Jang, Huiwon and Jeong, Jongheon and Huang, Jonathan and Shin, Jinwoo and Xie, Saining},
  booktitle={International Conference on Learning Representations},
  year={2025}
}

@inproceedings{leng2025repae,
  title={{REPA-E}: Unlocking {VAE} for End-to-End Tuning of Latent Diffusion Transformers},
  author={Leng, Xingjian and Singh, Jaskirat and Hou, Yunzhong and Xing, Zhenchang and Xie, Saining and Zheng, Liang},
  booktitle={Proceedings of the IEEE/CVF International Conference on Computer Vision},
  pages={18262--18272},
  year={2025}
}

@inproceedings{wang2025repa,
  title={{REPA} Works Until It Doesn't: Early-Stopped, Holistic Alignment Supercharges Diffusion Training},
  author={Wang, Ziqiao and Zhao, Wangbo and Zhou, Yuhao and Li, Zekai and Liang, Zhiyuan and Shi, Mingjia and Zhao, Xuanlei and Zhou, Pengfei and Zhang, Kaipeng and Wang, Zhangyang and Wang, Kai and You, Yang},
  booktitle={Advances in Neural Information Processing Systems},
  year={2025}
}

@article{singh2025what,
  title={What Matters for Representation Alignment: Global Information or Spatial Structure?},
  author={Singh, Jaskirat and Leng, Xingjian and Wu, Zongze and Zheng, Liang and Zhang, Richard and Shechtman, Eli and Xie, Saining},
  journal={arXiv preprint arXiv:2512.10794},
  year={2025}
}

@article{song2023consistencymodels,
      title={Consistency Models}, 
      author={Yang Song and Prafulla Dhariwal and Mark Chen and Ilya Sutskever},
      year={2023},
      journal={arXiv preprint arXiv:2303.01469},
}

@article{podell2023sdxl,
  title={Sdxl: Improving latent diffusion models for high-resolution image synthesis},
  author={Podell, Dustin and English, Zion and Lacey, Kyle and Blattmann, Andreas and Dockhorn, Tim and M{\"u}ller, Jonas and Penna, Joe and Rombach, Robin},
  journal={arXiv preprint arXiv:2307.01952},
  year={2023}
}

@inproceedings{lin2014microsoft,
  title={Microsoft coco: Common objects in context},
  author={Lin, Tsung-Yi and Maire, Michael and Belongie, Serge and Hays, James and Perona, Pietro and Ramanan, Deva and Doll{\'a}r, Piotr and Zitnick, C Lawrence},
  booktitle={European conference on computer vision},
  pages={740--755},
  year={2014},
  organization={Springer}
}

@article{lin2024sdxl,
  title={Sdxl-lightning: Progressive adversarial diffusion distillation},
  author={Lin, Shanchuan and Wang, Anran and Yang, Xiao},
  journal={arXiv preprint arXiv:2402.13929},
  year={2024}
}

@article{luo2023lcm,
  title={Lcm-lora: A universal stable-diffusion acceleration module},
  author={Luo, Simian and Tan, Yiqin and Patil, Suraj and Gu, Daniel and Von Platen, Patrick and Passos, Apolin{\~A}{\k{A}}rio and Huang, Longbo and Li, Jian and Zhao, Hang},
  journal={arXiv preprint arXiv:2311.05556},
  year={2023}
}
\bibliographystyle{plainnat}

%%%%%%%%%%%%%%%%%%%%%%%%%%%%%%%%%%%%%%%%%%%%%%%%%%%%%%%%%%%%

\newpage

\appendix

\section{Training Hyperparameters}\label{appendix: hyperparameters}

Table~\ref{tab:sdxl-imagenet-hyperparameters} summarizes the hyperparameters used for training on SDXL and ImageNet-$64\times 64$.

\begin{table*}[ht]
\centering
\scriptsize
\setlength{\tabcolsep}{4pt}
\renewcommand{\arraystretch}{1.05}
\caption{Hyperparameter settings for the SDXL and ImageNet-$64\times 64$ experiments.}
\label{tab:sdxl-imagenet-hyperparameters}
\begin{tabular}{@{}>{\raggedright\arraybackslash}p{0.22\textwidth}
                >{\raggedright\arraybackslash}p{0.36\textwidth}
                >{\raggedright\arraybackslash}p{0.36\textwidth}@{}}
\toprule
\textbf{Setting} & \textbf{SDXL} & \textbf{ImageNet-$64$} \\
\midrule
Task &
Text-to-image generation &
Class-conditional image generation \\
Training data &
LAION prompts with SDXL VAE latents &
ImageNet-$64$ training images \\
Teacher sampling for positives &
$50$ denoising steps, classifier-free guidance scale $6.0$ &
positives are real ImageNet samples from the same class \\
Generated samples per condition &
$4$ samples per prompt &
$4$ samples per class label \\
Positive / anchor samples per condition &
$4$ positive samples per prompt &
$4$ positives per class \\
Feature pooling &
Average pooling with pool size $4$ &
Average pooling with pool size $4$ \\
Drift radius set &
$\{0.02, 0.05, 0.2\}$ &
$\{0.02, 0.05, 0.1, 0.2\}$ \\
Coverage loss &
Anchor-margin coverage &
Anchor-margin coverage \\
Coverage weight &
$\lambda_{\mathrm{cov}}=1.0$ &
$\lambda_{\mathrm{cov}}=1.0$ \\
Coverage temperature &
$\tau=1.0$ &
$\tau=1.0$ \\
Anchor-margin scale &
$\alpha=0.5$ &
$\alpha=0.5$ \\
Feature-noise level &
0.1 &
0.1 \\
Generator learning rate &
$1 \times 10^{-6}$ &
$2 \times 10^{-6}$ \\
Optimizer &
AdamW, weight decay $0.01$ &
AdamW, weight decay $0.01$ \\
Learning-rate schedule &
Constant schedule with $500$ warmup steps &
Constant schedule with $500$ warmup steps \\
Precision &
FP16 &
FP16 \\
Gradient clipping &
Maximum norm $2.0$ &
Maximum norm $10.0$ \\
\bottomrule
\end{tabular}
\end{table*}

\section{Additional ImageNet Ablations}
\label{app:imagenet-ablations}

We provide additional ablations on class-conditional ImageNet-$64 \times 64$ to complement the main ImageNet results. Unless otherwise specified, all experiments use one step sampling and report the best FID observed no later than each update budget. Reporting budgeted best FID allows us to compare both early-stage optimization and later-stage performance.

\paragraph{Drift radii.}
Table~\ref{tab:imagenet64-drift-radius-budget-ablation} studies the kernel radius set used in the drifting field. The three tested radius sets produce similar performance, indicating that TFD is not overly sensitive to small changes in the kernel scale. The set $\{0.02,0.05,0.2\}$ gives the best final FID in this diagnostic comparison, supporting the use of averaged multi-radius drifting.
\begin{table}[h]
    \centering
    \caption{Drift-radius ablation on class-conditional ImageNet-$64 \times 64$. All rows use one step sampling, drift feature levels Enc-11 + Bottleneck + Dec-7, anchor-margin weight $\lambda_{\mathrm{anchor}}=1.0$, anchor-margin coefficient $\alpha=0.5$, and feature-noise at 0.5. Each entry reports the best FID observed no later than the specified update budget.}
    \label{tab:imagenet64-drift-radius-budget-ablation}
    \begin{tabular}{lcccc}
        \toprule
        Drift radii & $50$K & $100$K & $150$K & $200$K \\
        \midrule
        $\{0.02,0.05,0.1, 0.2\}$ & 2.172 & 1.875 & \textbf{1.695} & \textbf{1.624} \\
        $\{0.02,0.05,0.2\}$ & 2.183 & \textbf{1.874} & 1.724 & 1.648 \\
        $\{0.03,0.08,0.2\}$ & \textbf{2.130} & 1.880 & 1.728 & 1.703 \\
        $\{0.05,0.1,0.3\}$  & 2.160 & 1.909 & 1.742 & 1.698 \\
        \bottomrule
    \end{tabular}
\end{table}

\paragraph{Anchor-margin weight.}
Table~\ref{tab:imagenet64-anchor-weight-budget-ablation} studies the effect of the anchor-margin weight. The FID differences are modest, but nonzero anchor-margin weights improve the later-stage budgeted FID, with $\lambda_{\mathrm{anchor}}=1.0$ giving the best result before $200$K updates. Together with the diversity visualization in Figure~\ref{fig:anchor-margin-ablation}, this suggests that the anchor-margin term mainly improves coverage and reduces missing-mode behavior while maintaining competitive image quality.

\begin{table}[h]
    \centering
    \caption{Anchor-margin weight ablation on class-conditional ImageNet-$64 \times 64$. All rows use one step sampling, drift feature levels Enc-11 + Bottleneck + Dec-7, drift radii $\{0.02,0.05,0.2\}$, and feature-noise at 0.5. Each entry reports the best FID observed no later than the specified update budget.}
    \label{tab:imagenet64-anchor-weight-budget-ablation}
    \begin{tabular}{ccccc}
        \toprule
        $\lambda_{\mathrm{anchor}}$ & $50$K & $100$K & $150$K & $200$K \\
        \midrule
        $1.0$ & \textbf{2.194} & \textbf{1.819} &1.756 & \textbf{1.669} \\
        $0.5$ & 2.217 & 1.839 & \textbf{1.724} & 1.672 \\
        $0.0$ & 2.276 & 1.864 & 1.746 & 1.688 \\
        \bottomrule
    \end{tabular}
\end{table}

\begin{figure}[h]
    \centering
    \includegraphics[width=\linewidth]{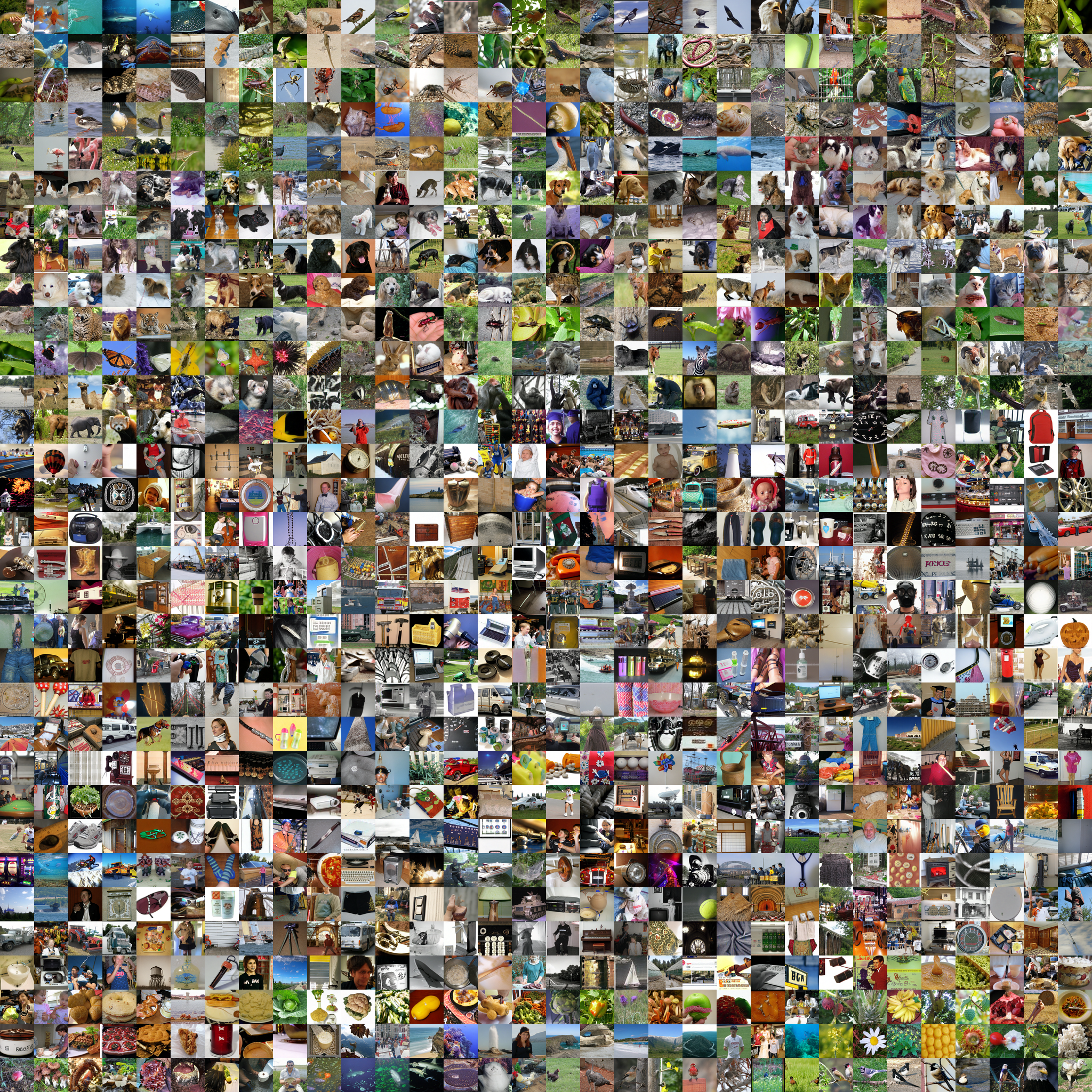}
    \caption{One step samples from our generator trained on ImageNet-$64\times64$}
    \label{fig:one step }
\end{figure}

\paragraph{Generator learning rate.}
Table~\ref{tab:imagenet64-generator-lr-budget-ablation} reports a diagnostic learning-rate comparison. The default learning rate $2\times 10^{-6}$ gives the strongest overall result among the available runs, while $5\times 10^{-6}$ is less stable and yields worse final FID. Since these runs are not a fully controlled sweep over all other hyperparameters, we use this comparison only as an appendix diagnostic and keep the main text focused on the fully matched ablations above.

\begin{table}[h]
    \centering
    \caption{Diagnostic generator learning-rate comparison on class-conditional ImageNet-$64 \times 64$. All rows use one step  sampling and the five-level feature configuration Enc-6 + Enc-11 + Bottleneck + Dec-7 + Dec-12. Each entry reports the best FID observed no later than the specified update budget.}
    \label{tab:imagenet64-generator-lr-budget-ablation}
    \begin{tabular}{cclcc}
        \toprule
        Generator LR   & $50$K & $100$K & $150$K & $200$K \\
        \midrule
        $1\times 10^{-6}$
        & 2.414 & 1.942 & 1.779 & 1.667 \\
        $2\times 10^{-6}$ 
        & \textbf{2.051} & \textbf{1.697} & \textbf{1.644} & \textbf{1.583} \\
        $5\times 10^{-6}$
        & 2.145 & 2.093 & 2.013 & 1.973 \\
        \bottomrule
    \end{tabular}
\end{table}

\begin{figure}
    \centering
    \includegraphics[width=\linewidth]{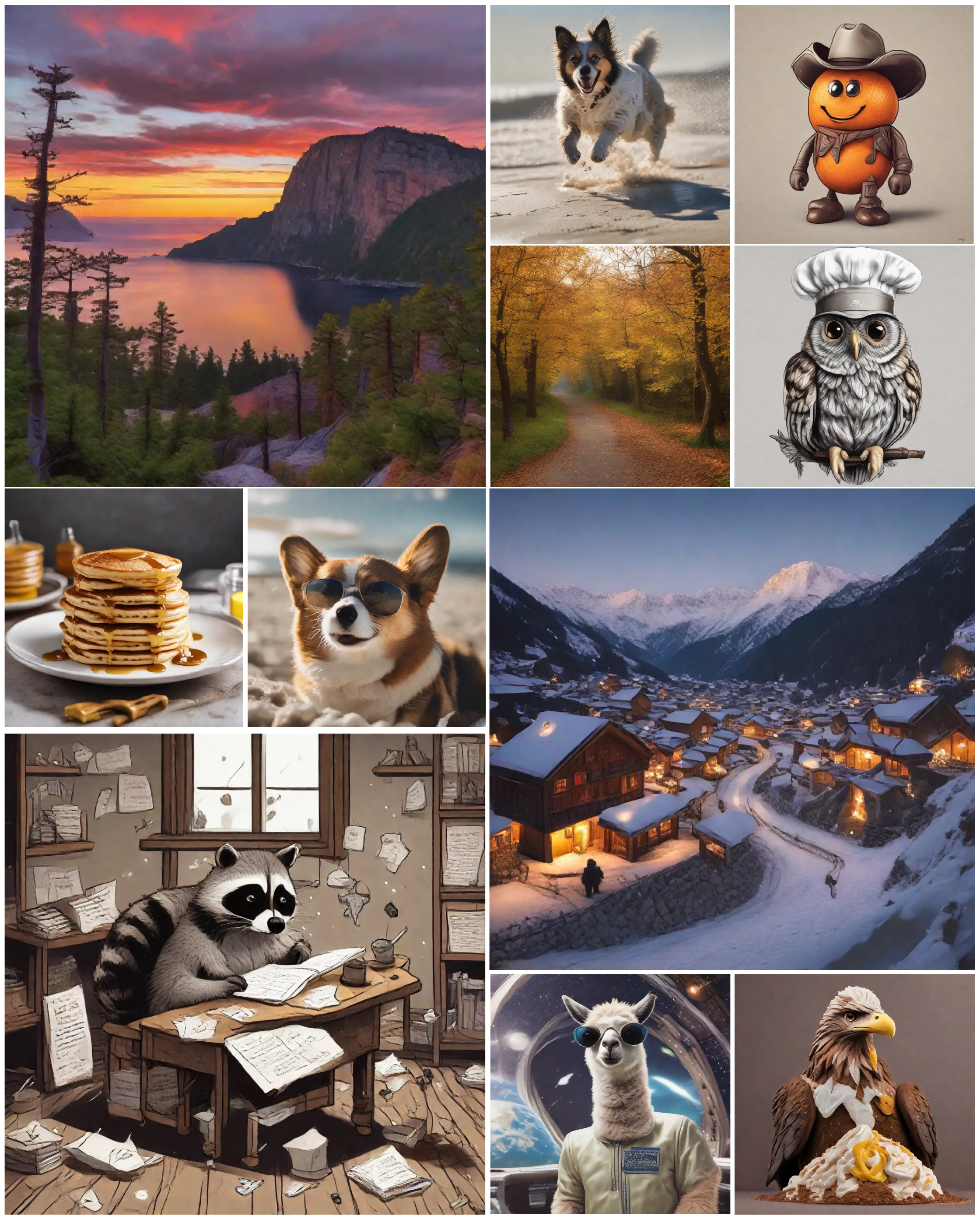}
    \caption{$1024\times1024$ samples produced by our \textbf{one step } generator distilled from SDXL.}
    \label{fig:sdxl}
\end{figure}

\section{Additional SDXL Ablations}
\label{app: sdxl abl}

\paragraph{Feature layers.}
Table~\ref{tab:sdxl-feature-layer-ablation} shows that the choice of SDXL U-Net feature layers has a substantial impact on training efficiency and final sample quality. Among the tested configurations, using the second down block, the middle block, and the first up block consistently achieves the best FID at all checkpoints. Replacing the second down block with the first residual layer of the first down block leads to noticeably worse convergence, especially at 2K and 3K updates, suggesting that the second down block provides more effective intermediate representations in this setting. In contrast, adding features from the second up block significantly degrades performance, indicating that simply increasing the number of feature layers does not necessarily improve generation quality and may introduce noisy or less transferable supervision signals.

\begin{table*}[t]
\centering
\caption{Controlled SDXL feature-layer ablation. Rows vary only the feature layers, while all other hyperparameters are kept fixed. {D}/{U} denote SDXL U-Net down/up blocks, {Mid} denotes the middle block, {Res} denotes a residual layer, and {Attn} denotes an attention layer. Early checkpoints report best-so-far FID at the specified training update.}
\label{tab:sdxl-feature-layer-ablation}
\begin{tabular}{cccc}
\toprule
\textbf{Feature layers} & \textbf{1K FID $\downarrow$} & \textbf{2K FID $\downarrow$} & \textbf{3K FID $\downarrow$} \\
\midrule
        D2 + Mid + U0 & \textbf{134.94} & \textbf{53.64} & \textbf{37.03}  \\
        D1-Res1 + Mid + U0 & 137.64 & 95.11 & 58.98 \\
        D2 + Mid + U0 + U1 & 267.34 & 218.81 & 142.28  \\
        D1-Attn1 + Mid + U1-Attn0 & 272.26 & 220.08 & 155.73 \\
\bottomrule
\end{tabular}
\end{table*}

\section{ImageNet Visual Results}
Figure~\ref{fig:one step } shows qualitative samples from the one step  ImageNet-$64\times64$ generator. Across diverse ImageNet classes, the generated images preserve recognizable class semantics and coherent object structure while requiring only a single forward pass at inference time. The samples also exhibit variation in pose, color, texture, and background composition, suggesting that the teacher-feature drifting objective does not merely collapse to a small set of class-specific prototypes. These visual results are consistent with the quantitative FID results in the main text and the appendix ablations, indicating that TFD can produce diverse and visually plausible class-conditional samples with a substantially simplified one step  distillation pipeline.

\section{SDXL Visual Results}
Figure~\ref{fig:sdxl} presents additional 1024$\times$1024 samples generated by our one step  SDXL student. The results show that TFD can preserve both global scene structure and local visual details at high resolution, covering diverse content types such as natural landscapes, animals, stylized characters, food, indoor scenes, and complex outdoor environments. Despite using only a single generator forward pass, the samples exhibit coherent composition, realistic textures, and strong visual diversity, further supporting the effectiveness of teacher-feature drifting for high-resolution text-to-image distillation.

\section{Broader Impacts}

TFD improves the efficiency of pretrained diffusion models by enabling competitive one-step generation with a simpler distillation pipeline. However, since TFD relies on pretrained teachers and preserves their semantic representation geometry, it may also inherit their biases, failure modes, and risks of misuse, including misleading synthetic content or copyright- and privacy-sensitive generation.

%%%%%%%%%%%%%%%%%%%%%%%%%%%%%%%%%%%%%%%%%%%%%%%%%%%%%%%%%%%%

\end{document}